# Asynchronous Forward Bounding for Distributed COPs

**Amir Gershman** AMIRGER@CS.BGU.AC.IL
**Amnon Meisels** AM@CS.BGU.AC.IL
**Roie Zivan** ZIVANR@CS.BGU.AC.IL
*Department of Computer Science,*
*Ben-Gurion University of the Negev,*
*Beer-Sheva, 84-105, Israel*

## Abstract

A new search algorithm for solving distributed constraint optimization problems (DisCOPs) is presented. Agents assign variables sequentially and compute bounds on partial assignments asynchronously. The asynchronous bounds computation is based on the propagation of partial assignments. The asynchronous forward-bounding algorithm (AFB) is a distributed optimization search algorithm that keeps one consistent partial assignment at all times. The algorithm is described in detail and its correctness proven. Experimental evaluation shows that AFB outperforms synchronous branch and bound by many orders of magnitude, and produces a phase transition as the tightness of the problem increases. This is an analogous effect to the phase transition that has been observed when local consistency maintenance is applied to MaxCSPs. The AFB algorithm is further enhanced by the addition of a backjumping mechanism, resulting in the *AFB-BJ* algorithm. Distributed backjumping is based on accumulated information on bounds of all values and on processing concurrently a queue of candidate goals for the next move back. The AFB-BJ algorithm is compared experimentally to other DisCOP algorithms (ADOPT, DPOP, OptAPO) and is shown to be a very efficient algorithm for DisCOPs.

## 1. Introduction

The Distributed Constraint Optimization Problem (*DisCOP*) is a general framework for distributed problem solving that has a wide range of applications in Multi-Agent Systems and has generated significant interest from researchers (Modi, Shen, Tambe, & Yokoo, 2005; Zhang, Xing, Wang, & Wittenburg, 2005; Petcu & Faltings, 2005a; Mailler & Lesser, 2004; Ali, Koenig, & Tambe, 2005; Silaghi & Yokoo, 2006). DisCOPs are composed of agents, each holding one or more variables. Each variable has a domain of possible value assignments. Constraints among variables (possibly held by different agents) assign costs to combinations of value assignments. Agents assign values to their variables and communicate with each other, attempting to generate a solution that is globally optimal with respect to the costs of the constraints (Modi et al., 2005; Petcu & Faltings, 2004).

There is a wide scope of motivation for research on DisCOP, since distributed COPs are an elegant model for many every day combinatorial problems that are distributed by nature. Take for example a large hospital that is composed of many wards. Each ward constructs a weekly timetable assigning its nurses to shifts. The construction of a weekly timetable involves solving a constraint optimization problem for each ward. Some of the nurses in every ward are qualified to work in the *Emergency Room*. Hospital regulations require a certain number of qualified nurses (e.g. for Emergency Room) in each shift. This imposes constraints among the timetables of different wards and generates a complex Distributed COP (Solotorevsky, Gudes, & Meisels, 1996).

Another example is the sensor networks tracking problem (Zhang, Xing, Wang, & Wittenburg, 2003; Zhang et al., 2005), in which the task is to assign sensors to tracking targets, such that the maximal number of the targets will be tracked by the sensor collection. This too can be solved using the DisCOP model.

DisCOP modeling can also solve problems like log based reconciliation (Chong & Hamadi, 2006), in which copies of a data base exist in several physical locations. Users perform actions on these data base copies, each user on its own local copy. The actions cause the data base to change, so only initially all copies are identical, but later actions change some of them and they are no longer identical. Logs of all user actions are kept. The problem is how to merge these logs, into a single log that keeps as many of the actions as possible. It is not always possible to keep all local logs intact, since actions are constrained with other actions (for example you can not reconcile the deletion of an item from the database and a later print or update of it).

DisCOPs represent real life problems that cannot or should not be solved centrally for several reasons, among them are lack of autonomy, single point of failure and privacy of agents. In the hospital wards example, wards want to maintain a degree of autonomy over their local problems involving the constraints of every single nurse. In the sensor example, the sensors have a very small memory and computing power and therefore cannot solve the problem in a centralized fashion. In the database example, centralization is possible, but issues such as network bottleneck, computing power and single point of failure encourage looking for a distributed solution.

The present paper proposes a new distributed search algorithm for *DisCOP*s, *Asynchronous Forward-Bounding (AFB)*. In the *AFB* algorithm agents assign their variables and generate a partial solution sequentially. The innovation of the proposed algorithm lies in propagating partial solutions asynchronously. Propagation of partial solutions enables asynchronous updating of bounds on their cost, and early detection of a need to backtrack, hence the algorithm's name *AFB*. This form of propagating bounds asynchronously turns out to generate a very efficient form of concurrent computation by all the participating agents. More efficient than algorithms that use asynchronous assignment processes, especially on hard instances of DisCOPs.

The overall framework of the *AFB* algorithm is based on a *Branch and Bound* scheme. Agents extend a partial solution as long as the lower bound on its cost does not exceed the global bound, which is the cost of the best solution found so far. In the proposed *AFB* algorithm, the state of the search process is represented by a data structure called *Current Partial Assignment (CPA)*. The CPA starts empty at some initializing agent that records its assignments on it and sends it to the next agent. The cost of a CPA is the sum on the costs of constraints it includes. Besides the current assignment cost, the agents maintain on a CPA a *lower bound* which is updated according to information they receive from yet unassigned agents. Each agent which receives the CPA, adds assignments of its local variables to the partial assignment on the received CPA, if an assignment with a lower bound smaller than the current global upper bound can be found. Otherwise, it backtracks by sending the CPA to a former agent to revise its assignment.

An agent that succeeds to extend the assignment on the CPA sends forward copies of the updated CPA, requesting all unassigned agents to compute lower bound estimations on the cost of the partial assignment. The assigning agent will receive these estimations asynchronously over time and use them to update the lower bound of the CPA.

Gathering updated lower bounds from future assigning agents, may enable an agent to discover that the lower bound of the CPA it sent forward is higher than the current upper bound (i.e. inconsistent). This discovery triggers the creation of a new CPA which is a copy of the CPA it sent forward. The agent resumes the search by trying to replace its inconsistent assignment. The time stamp mechanism proposed by Nguyen, Sam-Hroud, and Faltings (2004) and used by Meisels and Zivan (2007) is used by agents to determine the most updated CPA and to discard obsolete CPAs.

The concurrency of the *AFB* algorithm is achieved by the fact that forward-bounding is performed concurrently and asynchronously by all agents. This form of asynchronicity is similar to that employed by the Asynchronous Forward-Checking (*AFC*) algorithm for distributed constraint satisfaction problems (DisCSPs) (Meisels & Zivan, 2006; Meseguer & Jimenez, 2000). When *AFB* is enhanced with backjumping (Zivan & Meisels, 2007), the resulting algorithm performs concurrently distributed forward bounding and backjumping and prunes the search space of DisCOPs very efficiently. This is demonstrated by the extensive experimental evaluation in Section 6 where $AFB$ demonstrates a phase transition on randomly generated DisCOPs (Larrosa & Schiex, 2004). The extensive evaluation includes comparisons of the performance of $AFB$ to that of the best DisCOP search algorithms. These include asynchronous branch and bound like ADOPT (Modi et al., 2005), as well as algorithms that are based on other principles - DPOP (Petcu & Faltings, 2005a) that uses two passes on a pseudo-tree and $Opt_APO$,that divides the DisCOP into sub-problems (Mailler & Lesser, 2004).

The plan of the paper is as follows. Distributed Constraint Optimization are presented in Section 2. In Section 3, the $AFB$ algorithm in full details is presented. In Section 4 a version of the $AFB$ algorithm which is enhanced with conflict directed backjumping ($CBJ$) is presented. A correctness proof of the $AFB$ algorithm is presented in Section 5. In Section 6 an extensive empirical evaluation of the $AFB$ algorithm is presented. $AFB$ is compared with the state of the art DisCOP algorithms, $ADOPT$ which like $AFB$ does not include centralization of the problem's data and $DPOP$ and $Opt_APO$ (Petcu & Faltings, 2005a; Mailler & Lesser, 2004), which are based on very different principles. Our Conclusions are presented in Section 7.

## 2. Distributed Constraint Optimization

Formally, a $DisCOP$ is a tuple $< \mathcal{A}, \mathcal{X}, \mathcal{D}, \mathcal{R} >$. $\mathcal{A}$ is a finite set of agents $A_1, A_2, ..., A_n$. $\mathcal{X}$ is a finite set of variables $X_1$,$X_2$,...,$X_m$. Each variable is held by a single agent (an agent may hold more than one variable). $\mathcal{D}$ is a set of domains $D_1$, $D_2$,...,$D_m$. Each domain $D_i$ contains the finite set of values which can be assigned to variable $X_i$. $\mathcal{R}$ is a set of relations (constraints). Each constraint $C \in \mathcal{R}$ defines a none-negative *cost* for every possible value combination of a set of variables, and is of the form $C : D_{i_1} \times D_{i_2} \times \ldots \times D_{i_k} \rightarrow \mathbb{R}^+ \cup \{0\}$. A *binary constraint* refers to exactly two variables and is of the form $C_{ij} : D_i \times D_j \rightarrow \mathbb{R}^+ \cup \{0\}$. A *binary DisCOP* is a DisCOP in which all constraints are binary. An *assignment* (or a label) is a pair including a variable, and a value from that variable's domain. A *partial assignment* (PA) is a set of assignments, in which each variable appears at most once. *vars(PA)* is the set of all variables that appear in PA, $vars(PA) = \{X_i \mid \exists a \in D_i \wedge (X_i, a) \in PA\}$. A constraint $C \in \mathcal{R}$ of the form $C : D_{i_1} \times D_{i_2} \times \ldots \times D_{i_k} \rightarrow \mathbb{R}^+ \cup \{0\}$ is *applicable* to PA if $X_{i_1}, X_{i_2}, \ldots, X_{i_k} \in vars(PA)$. The *cost of a partial assignment* PA is the sum of all applicable constraints to PA over the assignments in PA. A *full assignment* is a partial assignment that includes all the variables ($vars(PA) = \mathcal{X}$). The goal is to find a full assignment of minimal cost.

In this paper, we will assume each agent owns a single variable, and use the term "agent" and "variable" interchangeably, and assume agent $A_i$ holds variable $X_i$ (Modi et al., 2005; Petcu & Faltings, 2005a; Mailler & Lesser, 2004). We will assume that constraints are at most binary and the delay in delivering a message is finite (Yokoo, 2000a; Modi et al., 2005). Furthermore, we assume a static final order on the agents, known to all agents participating in the search process (Yokoo, 2000a). These assumptions are commonly used by DisCSP and DisCOP algorithms (Yokoo, 2000a; Modi et al., 2005).

| $C_{ij}$ | | |
|---|---|---|
| $X_i$ | $X_j$ | Cost |
| R | R | 2 |
| R | B | 1 |
| B | R | 1 |
| B | B | 3 |

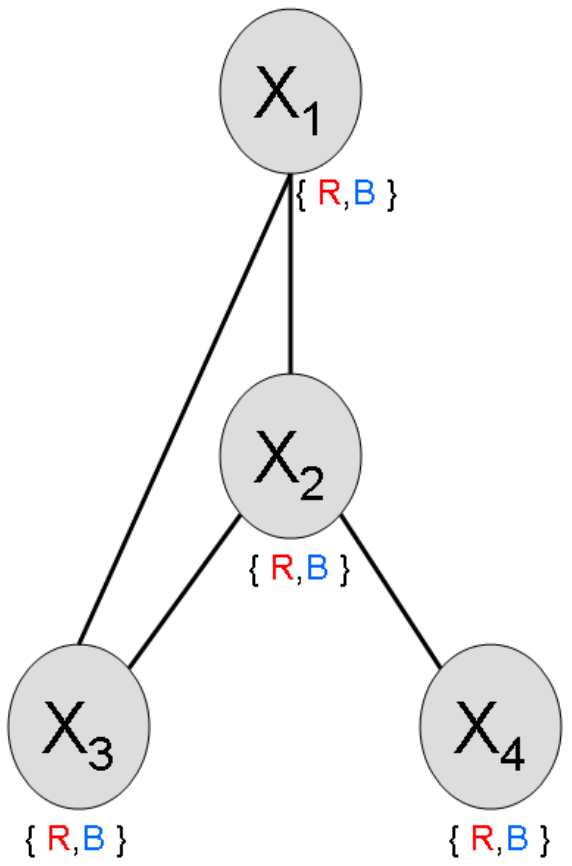

Figure 1: An example DisCOP. Each variable has two values R and B, all constraints are of the same form as shown in the table to the left.

**Example 1** *An example of a DisCOP is presented in figure 1. There are 4 variables, each variable is held by a different agent. The domains of all variables contain exactly the two values R and B. Lines between variables represent (binary) constraints. The cost of these constraints is shown in the table to the left. A partial assignment of* $\{(X_1, R)\}$ *has a cost of zero, since there is no constraint applicable to it. A partial assignment of* $\{(X_1, R), (X_4, R)\}$ *also has a cost of zero, since there is no constraint applicable to it. A partial assignment of* $\{(X_1, R), (X_2, R)\}$ *has a cost of two, due to the constraint* $C_{1,2}$*. A partial assignment of* $\{(X_1, R), (X_2, R), (X_3, B)\}$ *has a cost of four, due to the constraints* $C_{1,2}, C_{2,3}, C_{1,3}$*. One solution is* $\{(X_1, R), (X_2, B), (X_3, R), (X_4, R)\}$ *which has a cost of five. This is a solution since there is no other full assignment of lower cost.*

## 3. Asynchronous Forward Bounding

In the *AFB* algorithm a single most up-to-date current partial assignment is passed among the agents. Agents assign their variables only when they hold the up-to-date CPA.

The *CPA* is a unique message that is passed between agents, and carries the partial assignment that agents attempt to extend into a complete and optimal solution by assigning their variables on it. The CPA also carries the accumulated cost of constraints between all assignments it contains, as well as a unique time-stamp.

Due to the asynchronous nature of the algorithm, multiple CPAs may be present at any instant, however only a single CPA includes the most update to date partial assignment. This CPA has the highest timestamp.

Only one agent performs an assignment on a single CPA at any time. Copies of the CPA are sent forward and are concurrently processed by multiple agents. Each unassigned agent computes a lower bound on the cost of assigning a value to its variable, and sends this bound back to the agent which performed the assignment. The assigning agent uses these bounds to prune sub-spaces of the search-space which do not contain a full assignment with a cost lower than the best full assignment found so far. A total order among agents is assumed ($A_1$ is assumed to be the first agent in the order, and $A_n$ is assumed to be the last).

In more detail, every agent that adds its assignment to the CPA sends forward copies of the CPA, in messages we term *FB_CPA*, to all agents whose assignments are not yet on the CPA. An agent receiving an *FB_CPA* message computes a lower bound on the cost increment caused by adding an assignment to its variable. This estimated cost is sent back to the agent who sent the *FB_CPA* message via *FB_ESTIMATE* messages. The computation of this bound is detailed in section 3.1.

Notice that it is possible that the assigning agent already sent its CPA forward by the time the estimations are received. Should the estimations indicate that the CPA exceeds the bound, the agent will generate a new CPA, with a different local assignment (and a higher timestamp associated with it) and continue the search with this new CPA. The timestamping mechanism insures that the obsolete CPA will (eventually) be discarded regardless of its current location. The timestamp mechanism is described in section 3.3.

### 3.1 AFB - Computing the Lower Bound Estimation On Cost Increment

The computation of the lower bound on the cost increment caused by adding an assignment to the agent's local variable is done as follows.

Denote by $\mathbf{cost}((i, v), (j, u))$ the cost of assigning $A_i = v$ and $A_j = u$. For each agent $A_i$ and each value in its domain $v \in D_i$, we denote the minimal cost of the assignment ($i$,$v$) incurred by an agent $A_j$ by $h_j(v) = min_{u \in D_j}(cost((i, v), (j, u)))$. We define $h(v)$, the total cost of assigning the value $v$, to be the sum of $h_j(v)$ over all $j > i$. Intuitively, $h(v)$ is a lower bound on the cost of constraints involving the assignment $A_i = v$ and all agents $A_j$ such that $j > i$. Note that this bound can be computed once per agent, since it is independent of the assignments of higher priority agents.

An agent $A_i$, which receives an $FB_CPA$ message, can compute for every $v \in D_i$ both the cost increment of assigning $v$ as its value, i.e. the sum of the cost that $v$ has with the assignments included in the $CPA$, and $h(v)$. The sum of these, is denoted by $f(v)$. The lowest calculated $f(v)$ among all values $v \in D_i$ is chosen to be the lower bound estimation on the cost increment by agent $A_i$.

Figure 2 presents a constraint network. Large ovals represent variables while small circles represent values. In the presented constraint network, $A_1$ already assigned the value $v_1$ and $A_2, A_3, A_4$ are unassigned. Let us assume that the cost of every constraint is one. The cost of $v_3$ will increase by one due to its constraint with the current assignment thus $f(v_3) = 1$. Since $v_4$ is constrained with both $v_8$ and $v_9$, assigning this value will trigger a cost increment when $A_4$ performs an assignment. Therefore $h(v_4) = 1$ is an admissible lower bound of the cost of the constraints between this value and lower priority agents. Since $v_4$ does not conflict with assignments on the CPA, $f(v_4) = 1$ as well. $f(v_5) = 3$ because this assignment conflicts with the assignment on the CPA and in addition conflicts with all the values of the two remaining agents.

Since $h(v)$ takes into account only constraints of $A_i$ with lower priority agents ($A_j$ s.t. $j > i$), unassigned lower priority agents do not need to estimate their cost of constraints with $A_i$. Therefore, these estimations can be accumulated and summed up by the agent which initiated the forward bounding process to compute a lower bound on the cost of a complete assignment extended from the CPA.

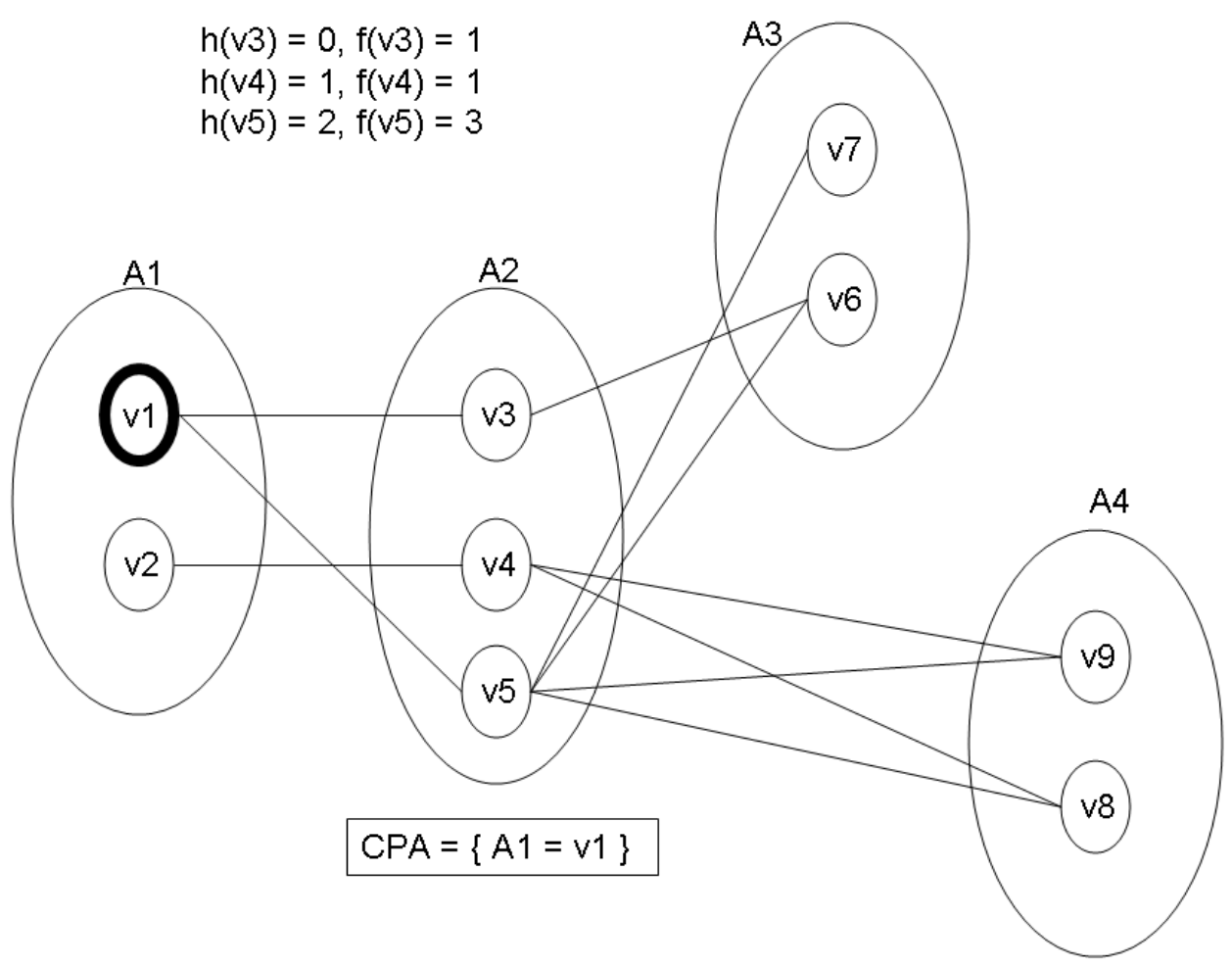

Figure 2: A simple DisCOP, demonstration

More formally we can define:

**Definition 1** ***CPA*** *is the current partial assignment, containing the assignments made by agents* $A_1, \ldots, A_{i-1}$.

Let us define the notions of past, local and future costs in definitions 2, 3 and 4.

**Definition 2** ***PC*** *(Past-Cost) is the added cost of assignments made by higher priority agents on the CPA (the costs incurred by agents* $A_1, \ldots, A_{i-1}$.

**Definition 3** ***LC(v)*** *(Local-Cost) is the cost incurred to the CPA if* $A_i$ *would assign the value* $v$ *and add it to the CPA. Therefore,*

$$LC(v) = \sum_{(A_j,w) \in CPA} cost((i,v),(j,w))$$

**Definition 4** ***FC(v)*** *(Future-Cost) is the sum of all lower bounds on cost increments caused by agents* $A_{i+1}, \ldots, A_n$ *for the CPA with the additional assignment of* $A_i = v$.

$$FC(v) = \sum_{j>i} min_{w \in D_j}(f(w)),\ s.t\ A_i = v\ added\ to\ CPA$$

The above definitions allow us to compute a lower bound on the cost of any full assignment extended from the CPA, and use this bound in order to prune parts of the search space. An agent ($A_i$) which receives the CPA, can question, what be its lower bound if it would be extended with an assignment of $A_i = v$. PC and LC(v) are both known to the agent, and FC(v) can be computed over time, by requesting future agents (lower priority agents) to compute their lower bounds and send them back to $A_i$. The sum PC + LC(v) + FC(v) composes this lower bound, and can be used to prune search spaces. This can happen when the agent knows that a full assignment was already found with cost lower than this sum, and therefore exploring this search-space would not lead to any better cost solutions.

Thus, asynchronous forward bounding enables agents an early detection of partial assignments that cannot be extended into complete assignments with cost smaller than the known upper bound, and initiate backtracks as early as possible.

### 3.2 AFB - Algorithm Description

The *AFB* algorithm is run on each of the agents in the *DisCOP*. Each agent first calls the procedure *init* and then responds to messages until it receives a $TERMINATE$ message. The algorithm is presented in Figure 3.1. The computation of bounds, and the time-stamping mechanism are not shown, as they are explained in the text.

In the initialization, each agent updates $B$ to be the cost of the best full assignment found so far and since no such assignment was found, it is set to infinity (line 1). Only the first agent ($A_1$) creates an empty CPA and then begins the search process by calling **assign_CPA** (lines 3-4), in order to find a value assignment for its variable.

An agent receiving a CPA (when received **CPA_MSG**), first makes sure it is relevant. The time stamp mechanism is used to determine the relevance of the CPA and will be explained in Section 3.3.

If the *CPA's* time-stamp reveals that it is not the most up to date CPA, the message is discarded. In such a case, the agent processing the message has already received a message implying that an assignment of some agent which has a higher priority than itself, has been changed. When the message is not discarded, the agent saves the received PA in its local CPA variable (line 7). Then, the agent checks that the received PA (without an assignment to its own variable) does not exceed the allowed cost $B$ (lines 8-10). If it does not exceed the bound, it tries to assign a value to its variable (or replace its existing assignment in case it has one already) by calling **assign_CPA** (line 13). If the bound is exceeded, a backtrack is initiated (line 11) and the CPA is sent to a higher priority agent, since the cost is already too high (even without an assignment to its variable).

Procedure **assign_CPA** attempts to find a value assignment, for the current agent, within the bounds of the current CPA. First, estimates related to prior assignments are cleared (line 19). Next, the agent attempts to assign every value in its domain it did not already try. If the CPA arrived without an assignment to its variable, it tries every value in its domain. Otherwise, the search for such a value is continued from the value following the last assigned value. The assigned value must be such that the sum of the cost of the CPA and the lower bound of the cost increment caused by the assignment will not exceed the upper bound $B$ (lines 20-22). If no such value is found, then the assignment of some higher priority agent must be altered, and so backtrack is called (line 23). Otherwise, the agent assigns the selected value on the CPA.

When the agent is the last agent ($A_n$), a complete assignment has been reached, with an accumulated cost lower than $B$, and it is broadcasted to all agents (line 27). This broadcast will inform the agents of the new bound for the cost of a full assignment, and cause them to update their upper bound $B$.

The agent holding the CPA ($A_n$) continues the search, by updating its bound $B$, and calling **assign_CPA** (line 29). The current value will not be picked by this call, since the *CPA's* cost with this assignment is now equal to B, and the procedure requires the cost to be lower than $B$. So the agent will continue the search, testing other values, and backtracking in case they do not lead to further improvement.

```
procedure init:
  B ← ∞
  if (A_i = A_1)
    generate_CPA()
    assign_CPA()

when received (FB_CPA, A_j, PA)
  f ← estimation based on the received PA.
  send (FB_ESTIMATE, f, PA, A_i) to A_j

when received (CPA_MSG, PA)
  CPA ← PA
  TempCPA ← PA
  if TempCPA contains an assignment to A_i, remove it
  if (TempCPA.cost ≥ B)
    backtrack()
  else
    assign_CPA()

when received (FB_ESTIMATE, estimate, PA , A_j)
  save estimate
  if ( CPA.cost + all saved estimates) ≥ B )
    assign_CPA()

when received (NEW_SOLUTION, PA)
  B_CPA ← PA
  B ← PA.cost

procedure assign_CPA:
  clear estimations
  if CPA contains an assignment A_i = w, remove it
  iterate (from last assigned value) over D_i until found
  v ∈ D_i s.t. CPA.cost + f(v) < B
  if no such value exists
    backtrack()
  else
    assign A_i = v
    if CPA is a full assignment
      broadcast (NEW_SOLUTION, CPA )
      B ← CPA.cost
      assign_CPA()
    else
      send(CPA_MSG, CPA) to A_{i+1}
      forall j > i
      send(FB_CPA, A_i, CPA) to A_j

procedure backtrack:
  clear estimates
  if (A_i = A_1)
    broadcast(TERMINATE)
  else
    send(CPA_MSG, CPA) to A_{i−1}
```

Figure 3: The procedures of the AFB Algorithm

When the agent holding the CPA is not the last agent (line 30), the CPA is sent forward to the next unassigned agent, for additional value assignment (line 31). Concurrently, forward bounding requests (i.e. **FB_CPA** messages) are sent to all lower priority agents (lines 32-33).

An Agent receiving a forward bounding request (when received **FB_CPA**) from agent $A_j$, again uses the time-stamp mechanism to ignore irrelevant messages. Only if the message is relevant, then the agent computes its estimate (lower bound) of the cost incurred by the lowest cost assignment to its variable (line 5). The exact computation of this estimation was described in Section 3.1 (it is the minimal $f(v)$ over all $v \in D_i$). This estimation is then attached to the message and sent back to the sender, as a **FB_ESTIMATE** message.

An agent receiving a bound estimation (when received **FB_ESTIMATE**) from a lower priority agent $A_j$ (in response to a forward bounding message) ignores it if it is an estimate to an already abandoned partial assignment (identified by using the time-stamp mechanism). Otherwise, it saves this estimate (line 14) and checks if this new estimate causes the current partial assignment to exceed the bound $B$ (line 15). In such a case, the agent calls $assign_CPA$ (line 16) in order to change its value assignment (or backtrack in case a valid assignment cannot be found).

The call to **backtrack** is made whenever the current agent cannot find a valid value (i.e. below the bound B). In such a case, the agent clears its saved estimates, and sends the CPA backwards to agent $A_{i-1}$ (line 38). If the agent is the first agent (nowhere to backtrack to), the terminate broadcast ends the search process in all agents (line 36). The algorithm then reports that the optimal solution has a cost of $B$, and the full assignment with such a cost is $B_CPA$.

### 3.3 The Time-Stamp Mechanism

As mentioned previously, AFB uses a time-stamp mechanism (Nguyen et al., 2004; Meisels & Zivan, 2007) to determine the relevance of the CPA. The requirements from this mechanism are that given two messages with two different partial assignments, it must determine which one of them is obsolete. An obsolete partial assignment is one that was abandoned by the search process because one of the assigned agents has changed its assignment. This requirement is accomplished by the time-stamping mechanism in the following way. Each agent keeps a local running-assignment counter. Whenever it performs an assignment it increments its local counter. Whenever it sends a message containing its assignment, the agent copies its current counter onto the message. Each message holds a vector containing the counters of the agents it passed through. The i-th element of the vector corresponds to $A_i$'s counter. This vector is in fact the time-stamp. A lexicographical comparison of two such vectors will reveal which time-stamp is more up-to-date.

Each agent saves a copy of what it knows to be the most up-to-date time-stamp. When receiving a new message with a newer time-stamp, the agent updates its local saved “latest” time-stamp. Suppose agent $A_i$ receives a message with a time-stamp that is lexicographically smaller than the locally saved “latest”, by comparing the first $i - 1$ elements of the vector. This means that the message was based on a combination of assignments which was already abandoned and this message is discarded. Only when the message's time-stamp in the first $i - 1$ elemental is equal or greater than the locally saved ”best” time-stamp is the message processed further.

The vector's counters might appear to require a lot of space, as the number of assignments can grow exponentially in the number of agents. However, if the agent ($A_i$) resets its local counter to zero each time the assignments of higher priority agents are altered, the counters will remain small (log of the size of the value domain), and the mechanism will remain correct.

### 3.4 AFB - Example Run

Suppose we run AFB on the DisCOP in figure 1. $X_1$ will create an empty CPA, assign its first value $R$ and pass the CPA to $X_2$. The CPA will travel from $X_2$, to $X_3$ and finally to $X_4$, with each agent assigning its first value ($R$) on it along the way until finally at $X_4$ we will have a full assignment with total accumulated cost of 8. This cost will be broadcasted to all agents (line 27 in figure 3.1) as the new upper bound (instead of infinity). Next, $X_4$ will call the $assign_CPA$ procedure (line 29). This call will result in a new assignment for $X_4$, with the value $B$, since the resulting full assignment will have a cost of only 7. This will cause another broadcast update of the upper bound and another call to $assign_CPA$. In this next call, $X_4$ will have an empty domain and be forced to backtrack the CPA to $X_3$. This CPA contains the assignments $X_1 = X_2 = X_3 = R$, with a total accumulated cost of 6 which is below the upper bound. Therefore $X_3$ will call its $assign_CPA$ (line 13). Examining its remaining values, $X_3$ explores the assignment of $B$ which will result in a CPA with a cost of 4 (line 21), which is below the current upper bound B. The CPA is sent to $X_4$ (line 31). $X_4$ calls the $assign_CPA$ procedure (line 13). The value $R$ will result in a CPA with a cost of 6, which is better than the upper bound $B$ of 7, and therefore is broadcasted (line 27). The next value, $B$, explored by $X_4$ results in a CPA with cost 5, which is also broadcasted. The CPA is sent backwards to $X_3$. $X_3$ has no more values to try, so it also backtracks the CPA, to $X_2$. $X_2$ assigns its next value, $B$, and sends the CPA to $X_3$. In addition $X_2$ also sends copies of the CPA in FB_CPA messages to $X_3$ and $X_4$ (line 33). If $X_3$ now receives this FB_CPA, it computes an estimation of 3 (because if $X_3$ is $R$ then it would increase this CPA's cost by 3 and if it were $B$ it would increase it by 4), and sends this information back to $X_2$ (line 6). Suppose $X_4$ also receives his $FB_CPA$, it then replies with an estimation of 1. While the CPA explores the sub-search in which $X_2 = B$ (passing between $X_3$ and $X_4$), these estimations arrive at $X_2$. $X_2$ saves these estimations and adds them up. This leads to the discovery that a backtrack is needed, since the CPA's cost is 1 (because $X_1 = R, X_2 = B$) with the additional estimations of 4 results in a sum equal to the upper bound $B$ (line 15). Therefore, $X_2$ abandons its assignment and attempts to assign its next value (calling $assign_CPA$ - line 16). Since $X_2$ has no values, this call results in a backtrack (line 23). The CPA sent from this backtrack has a higher timestamp value than the CPA previously sent forward by $X_2$, and the former CPA would eventually be discarded.

### 3.5 Discussion - Concurrency, Robustness, Privacy and Asynchronicity

At any point in time during the run of AFB, there is a single most-up-to-date CPA in the system. Each agent adds an assignment when it holds it, so assignments are performed sequentially. One might think that this would necessarily result in poor performance, as the search process does not try to take advantage of the existing multiple computational resources available to it. The concurrency of AFB comes from the use of the forward-bounding mechanism. While the CPA is held by one agent, many copies of it are sent forward, and a collection of agents compute concurrently lower bounds for that CPA. When the CPA advances to the next agent, again this process repeats, and so the unassigned agents are constantly kept working, either when they receive the CPA, or when they need to compute bounds for some other partial assignment.

This degree of asynchronicity is similar to that employed by the Asynchronous Forward-Checking *AFC* algorithm for DisCSPs (Meseguer & Jimenez, 2000; Meisels & Zivan, 2006). *AFC* performs a similar process in which the agents receive "forward-checking" messages by agents which performed assignments. The unassigned agents perform forward-checking (checking they have at least one value which is consistent with all previous assignments). In *AFB* these agents compute a lower bound on their local cost increment due to all assignments made by previous agents. Due to this similarity we named our algorithm Asynchronous Forward-Bounding.

AFB's approach is quite different from that used by asynchronous assignments algorithms such as *ADOPT* or *ABT* (Modi et al., 2005; Bessiere, Maestre, Brito, & Meseguer, 2005). In these algorithms the search process attempts to perform assignments concurrently by the collection of agents. Since many agents are assigning their variables simultaneously, there is a probability that must be handled by the algorithm, that the current agent's view of assignments made by other agents is incorrect. This is due to the fact that agents concurrently alter their assignments. The algorithm must be able to deal with this uncertainty.

A search process which performs assignments asynchronously may be expected to save time since agents need not wait for all assignments of past agents to reach them, as is done by a sequentially assigning algorithm. However, asynchronously assigning algorithms must also deal with inconsistencies caused by message delay. For example, if several higher priority agents change their assignments and only some of the messages are received (the others are delayed) computation performed will be based on this inconsistent agent view. This type of scenario, which has computation based on an inconsistent partial assignment, is completely avoided by sequentially assigning algorithms.

One variation of the AFB algorithm has agents which sent out FB-CPA messages, send these messages only to the subset of the target agents which have a direct constraint with the sending agent. This may be useful if the communication between agents is limited (agents may only communicate with agents with whom they have a direct conflict) and would keep the algorithm correct. This change may have two effects. First, less agents will return bounds to the sending agents. These bounds can be significant (greater than zero) since they take into account constraints with assignments of previous agents (which they may be conflicted with) and also constraints between the receiving agent and agents of lower priority (constraint between unassigned agents). Receiving less lower bounds would not invalidate the correctness of the algorithm but it may cause the search process to needlessly explore sub-spaces which could have been discovered to be dead-ends. Second, the detection of obsolete CPAs may be delayed since less agents receive a higher timestamp (which the FB-CPA may contain). The mechanism would remain correct since eventually another FB-CPA or the CPA itself would reach an agent which did not receive the FB-CPA, however this may take more time than a single ”cycle” of messages (in other words, more time than the travel time of a single message between two agents). The AFB algorithm was intentionally presented as an algorithm which sends out FB messages to all unassigned agents, since no constraint on communication between agents is assumed. In case such constraints exist, or one attempts to reduce the number of messages sent by the algorithm, this variation should be explored.

Privacy is considered one of the main motivations for solving problems distributively. The common model for distributed search algorithms on DisCSPs and DisCOPs enables assignments and *Nogoods* to be passed among agents (Yokoo, Ishida, Durfee, & Kuwabara, 1992; Yokoo, 2000b; Bessiere et al., 2005; Modi et al., 2005; Zivan & Meisels, 2006; Meisels & Zivan, 2007). $AFB$ follows the model proposed by Yokoo, sending assignments forward and bounds on partial assignments ($Nogoods$) backwards. An additional privacy drawback of $AFB$ is the fact that agents can learn about the assignments of non neighboring agents via CPAs which they receive from their neighbors. This problem can be easily solved in $AFB$ by a simple use of encryption. If every pair neighboring agents will share an encryption key, then an agent would be able to learn only the assignments of its neighbors when it receives a CPA. Such use of limited encryption in DisCOP algorithms was recently proposed for $DPOP$ by (Greenstadt, Grosz, & Smith, 2007).

If, due to privacy, the constraints are partially known so that between two constrained agents, only a part of the constraint is known to each of the constrained agents, then the bound computation mechanism must be adjusted in AFB. These type of constraints were discussed for DisCSP algorithms (Brito, Meisels, Meseguer, & Zivan, 2008). To the best of our knowledge, no DisCOP solver so far has handled such constraints. This remains an interesting possible extension to AFB as part of future work.

Robustness is another important aspect of a distributed search algorithm. We assumed that all messages are delivered in the order in which they are sent and no messages are lost. However if message passing is susceptible to losses or corruption of the data, AFB may not terminate (if, say, the CPA message is lost). It is also possible that the local data held by some agents will be corrupt (due to some mechanical failure for example). A solution would be to build a self-stabilizing algorithm. Self stabilization in distributed systems (Dijkstra, 1974) is the ability of a system to respond to transient failures by eventually reaching and maintaining a legal state. A self stabilizing version was shown for a simple DFS algorithm for DisCSPs (Collin, Dechter, & Katz, 1999). Based on that self-stabilizing DFS algorithm, a self-stabilizing version of DPOP was developed (Petcu & Faltings, 2005b). However these are the only self-stabilizing DisCSP/DisCOP solvers to the best of the authors' knowledge. Clearly, a more thorough study of robustness and self-stabilization is required for DisCOP algorithms.

To conclude, The *AFB* algorithm includes concurrent computation by multiple agents, without having to deal with the uncertainty that comes with asynchronous assignments. Each agent that receives a message containing a partial assignment knows with certainty that the given partial assignment is the one it was supposed to receive, and not a result of a network delay inconsistency. Therefore, *AFB* has both concurrent computation and the certainty of working with consistent partial assignments. This results in a much better performance on hard instances of random DisCOPs, as will be demonstrated in the empirical evaluation in section 6.

## 4. AFB with CBJ

In both centralized and distributed *CSP*s backjumping can be accomplished by maintaining data structures that allow an agent to deduce who is the latest agent (in the order in which assignments were made) whose changed assignment could possibly lead to a solution. Once such an agent is found, the assignments of all following agents are unmade and the search process "backjumps" to that agent (Prosser, 1993).

A similar process can be designed for branch and bound based solvers for *COP*s and *DisCOP*s. Consider a sequence of assignments by the agents $A_1, A_2, A_3, A_4, A_5$ where $A_5$ determined that none of its possible value assignments can lead to a full assignment with a cost lower than the cost of the best full assignment found so far. Clearly, $A_5$ must backtrack.

In chronological backtracking, the search process would simply return to the previous agent, namely $A_4$, and have it change its assignment. However, $A_5$ can sometimes determine that no value change of $A_4$ would suffice to reach a full assignment with a lower cost. Intuitively, $A_5$ can safely backjump to $A_3$, if it can compute a lower bound on the cost of a full assignment extended from the assignments of $A_1$, $A_2$ and $A_3$, and show that this bound is greater or equal to the cost of the best full assignment found so far. This is the intuitive basis of how backjumping can be added to *AFB*.

More formally, let us consider a scenario in which $A_i$ decides to backtrack, and the cost of the best full assignment found so far is $B$ (e.g. the upper bound of the current state of the search). The current partial assignment includes the assignments of agents $A_1, ..., A_{i-1}$.

**Definition 5** ***CPA[1..k]*** *is the set of assignments made by agents $A_1, \ldots, A_k$ in the current partial assignment. We define $CPA[1..0] = \{\}$.*

**Definition 6** ***FA[k]*** *is the set of all full assignments, which include all the assignments appearing in CPA[1..k]. In other words, this set contains all full assignments which can be extended from the assignments appearing in CPA[1..k]. Naturally, FA[0] is the set of all possible full assignments.*

On a backtrack, instead of simply backtracking to the previous agent, $A_i$ performs the following actions: It computes a lower bound on the cost of any full assignment in FA[i-2]. If this bound is smaller than $B$, it backtracks to $A_{i-1}$ just like it would do in chronological backtracking. However, if this bound is greater or equal to $B$, then backtracking to $A_{i-1}$ would do little good. No value change of $A_{i-1}$ alone could result in a full assignment of cost lower than $B$. As a result, $A_i$ knows it can safely backjump to $A_{i-2}$. It may be possible for $A_i$ to backjump even further, depending on the lower bound on the cost of any full assignment in FA[i-3]. If this bound is smaller than $B$, it backjumps to $A_{i-2}$. Otherwise, it knows it can safely backjump to $A_{i-3}$. Similar checks can be made about the necessity to backjump further.

The backjumping procedure relies on the computation of lower bounds for sets of full assignments (FA[k]). Next, we will show how can $A_i$ compute such lower bounds. Let us define the notions of past, local and future costs in definitions 7, 8 and 9.

**Definition 7** ***PC*** *(Past-Costs) is a vector of size n+1, in which the k-th element ($0 \leq k \leq n$) is equal to the cost of CPA[1..k].*

**Definition 8** ***LC(v)*** *(Local-Costs) is a vector of size $n+1$ computed by $A_i$ and held by it, in which the k-th element ($0 \leq k \leq n$) is*

$$LC(v)[k] = \sum_{(A_j,v_j) \in CPA \; s.t \; j \leq k} cost(A_i = v, A_j = v_j)$$

*Since the CPA held by $A_i$ only includes assignments of $A_1, \ldots, A_{i-1}$, then*

$$\forall j \geq i, LC(v)[i-1] = LC(v)[j]$$

*Intuitively, LC(v)[i] is the accumulated cost of the value $v$ of $A_i$, with respect to all assignments in CPA[1..i].*

**Definition 9** ***FC$_j$(v)*** *(Future-Costs) is a vector of size n+1, in which the k-th element ($0 \leq k \leq n$) contains a lower bound on the cost of assigning a value to $A_j$ with respect to the partial assignment CPA[1..k]. Assume this structure is held by agent $A_i$. If $k \geq i$ then CPA[1..k] contains the assignment $Ai = v$, but for $k < i$ the value $v$ of $A_i$ is irrelevant as it does not appear in CPA[1..k].*

The above vectors provide additive lower bounds on full assignments that start with the current CPA up to $k$, FA[k]. PC[k] is the exact cost of the first $k$ assignments, LC(v)[k] is the exact cost of the assignment $A_i = v$, and $\sum_{j>i} FC_j(v)[k]$ is a lower bound on the assignments of $A_{i+1}$, ..., $A_n$. Therefore, the sum

$$\textbf{FALB(v)[k]} = LC(v)[k] + PC[k] + \sum_{j>i} FC_j(v)[k]$$

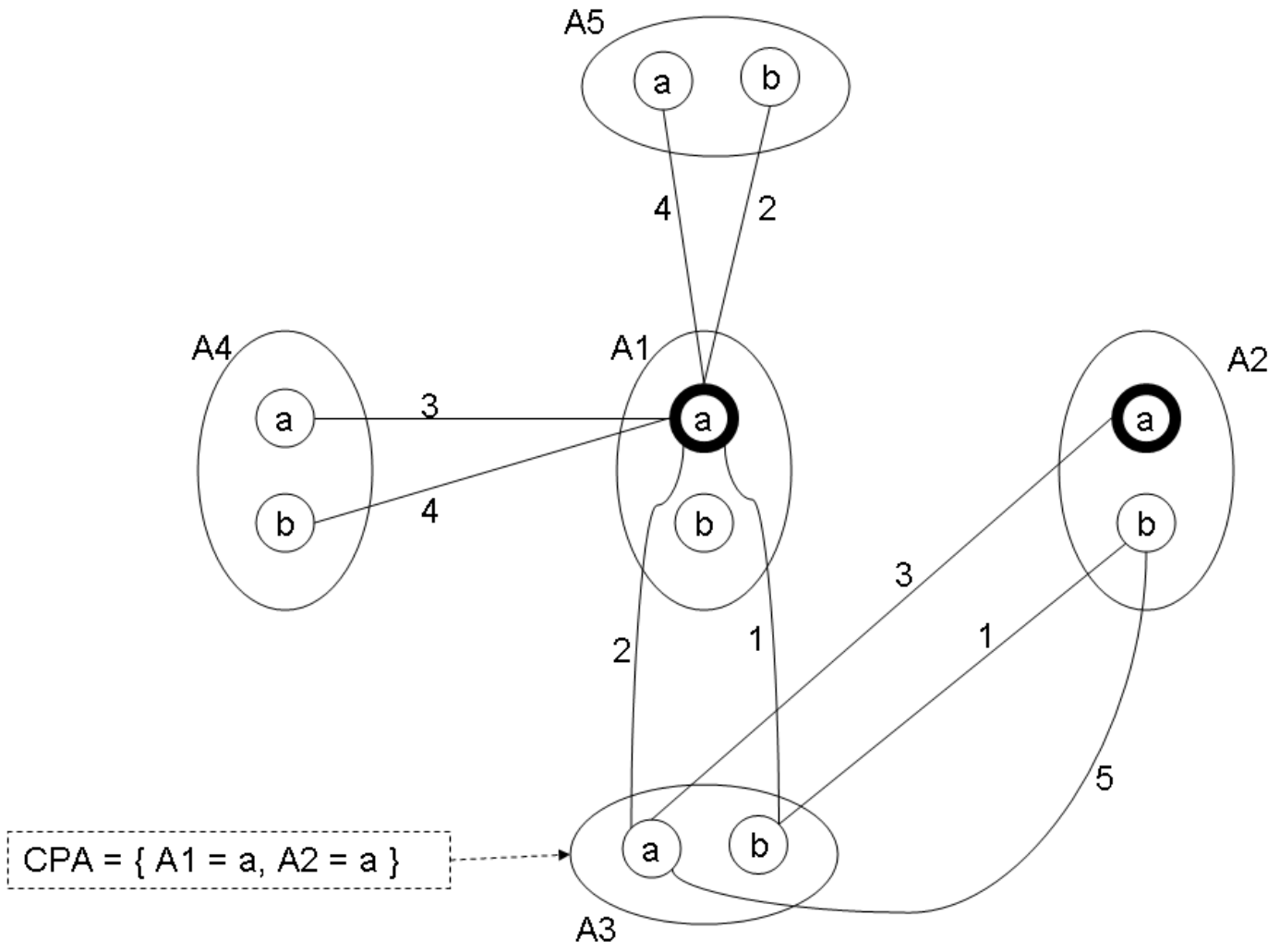

Figure 4: An example DisCOP

is a *Full Assignment Lower Bound* on the cost of any full assignment extended from CPA[1..k] in which $A_i = v$.

FA[k] contains all full assignments extended from CPA[1..k], and is not limited to assignments in which $A_i = v$. If we go over all FALB(v)[k], for all possible values $v \in D_i$ we produce a lower bound on any assignment in FA[k].

**Definition 10** ***FALB[k]*** $= \min_{v \in D_i}(FALB(v)[k])$.
*FALB[k] is a lower bound on the cost of any full assignment extended from CPA[1..k].*

In a distributed branch and bound algorithm, this bound is computed by $A_i$. PC - the cost of previous agents is sent along with their value assignment messages to $A_i$. LC(v) - the cost of assigning $v$ to $A_i$ can be computed by $A_i$. $A_i$ requests all agents ordered after it, $A_j$ ($j > i$), to compute FC$_j$ and send the results back to $A_i$. This is part of the already existing AFB mechanism for forward bounding.

In the *AFB* algorithm (Gershman, Meisels, & Zivan, 2007) $A_i$ already requests unassigned agents to compute lower bounds on the CPA and send back the results. The additional bounds needed for backjumping can be easily added to the existing *AFB* framework.

### 4.1 A Backjumping Example

To demonstrate the backjumping possibility, consider the DisCOP in Figure 4 (again, large ovals represent variables while small circles represent values). Let us assume that the search begins with $A_1$ assigning “a” as its value and sending the $CPA$ forward to $A_2$. $A_2, A_3, A_4$, and $A_5$ all assign the value “a” and we get a full assignment with cost 12. The search continues, and after fully exploring the sub-space in which $A_1 = a, A_2 = a$, the best assignment found is $A_1 = a, A_2 = a, A_3 = b, A_4 = a, A_5 = b$ with a total cost of $B$=6. Assume that $A_3$ is now holding the $CPA$ after receiving it from some future agent ($A_4$ or $A_5$). $A_3$ has exhausted its value domain and must backtrack. It computes:

$$FALB(a)[1] = PC[1] + LC(a)[1] + (FC_4(a)[1] + FC_5(a)[1])$$

$$= 0 + 2 + (3 + 2) = 7$$

$$FALB(b)[1] = PC[1] + LC(b)[1] + (FC_4(b)[1] + FC_5(b)[1])$$

$$= 0 + 1 + (3 + 2) = 6$$

$$FALB[1] = min(FALB(a)[1], FLAB(b)[1]) = 6$$

$FALB[1] \geq B$, therefore $A_3$ knows that any full assignment extended from $\{A_1 = a\}$ would cost at least 6. A full assignment with that cost was already discovered, so there is no need to explore the rest of this sub-space, and it can safely backjump the search process back to $A_1$, to change its value to "b". Backtracking to $A_2$ leaves the search process within the $\{A_1 = a\}$ sub-space, which $A_3$ knows cannot lead to a full assignment with a lower cost.

### 4.2 The AFB-BJ Algorithm

The *AFB-BJ* algorithm is run on each of the agents in the *DisCOP*. Each agent first calls the procedure *init* and then responds to messages until it receives a TERMINATE message. The algorithm is presented in figures 5 and 6. As in pure *AFB*, a timestamping mechanism is used on all messages.

The same timestamping mechanism used by AFB is used in AFB-BJ to determine which messages are relevant and which are obsolete. For simplicity we choose to omit the pseudo-code detailing the calculation of LC, PC, FC and FALB, as they were described in Section 4.1.

The algorithm starts by each agent calling **init** and then awaiting messages until termination. At first, each agent updates $B$ to be the cost of the best full assignment found so far and since no such assignment was found, it is set to infinity (line 1). Only the first agent ($A_1$) creates an empty CPA and then begins the search process by calling **assign_CPA** (lines 3-4), in order to find a value assignment for its variable.

An agent receiving a CPA (when received **CPA_MSG**), checks the time-stamp associated with it. An out of date $CPA$ is discarded. When the message is not discarded, the agent saves the received PA in its local CPA variable (line 7). In case the CPA was received from a higher priority agent, the estimations of future agents in $FC_j$ are no longer relevant and are discarded, and the domain values must be reordered by their updated cost (lines 9-11). Then, the agent attempts to assign its next value by calling **assign_CPA** (line 16) or to backtrack if needed (line 14).

Procedure **assign_CPA** attempts to find a value assignment, for the current agent. The assigned value must be such that the sum of the cost of the CPA and the lower bound of the cost increment caused by the assignment will not exceed the upper bound $B$ (lines 23). If no such value is found, then the assignment of some higher priority agent must be altered, so backtrack is called (line 25). When a full assignment is found which is better than the best full assignment known so far, it is broadcast to all agents (line 29). After succeeding to assign a value, the CPA is sent forward to the next unassigned agent (line 33). Concurrently, forward bounding requests (i.e. FB_CPA messages) are sent to all lower priority agents (lines 34-35).

An agent receiving a bound estimation (when received **FB_ESTIMATE**) from a lower priority agent $A_j$ (in response to a forward bounding message) ignores it if it is an estimate to an already abandoned partial assignment (identified using the time-stamp mechanism). Otherwise, it saves this estimate (line 17) and checks if this new estimate causes the current partial assignment to exceed the bound $B$ (line 18). In such a case, the agent calls $assign_CPA$ (line 19) in order to change its value assignment (or backtrack in case a valid assignment cannot be found).

procedure **init**:
1. $B \leftarrow \infty$
2. **if** ($A_i = A_1$)
3. $\quad generate_CPA()$
4. $\quad assign_CPA()$

when received (**FB_CPA**, $A_j$, $PA$)
5. $V \leftarrow$ estimation vector for each PA[1..k] ($0 \leq k \leq n$)
6. send ($FB_ESTIMATE$, $V$, $PA$, $A_i$) to $A_j$

when received (**CPA_MSG**, $PA$, $A_j$)
7. $CPA \leftarrow PA$
8. $TempCPA \leftarrow PA$
9. **if** ($j = i - 1$)
10. $\quad \forall j$ re-initialize $FC_j(v)$
11. $\quad$ reorder domain values $v \in D_i$ by LC(v)[i] (from low to high)
12. **if** ($TempCPA$ contains an assignment to $A_i$) remove it
13. **if** ($TempCPA.cost \geq B$)
14. $\quad$ backtrack()
15. **else**
16. $\quad assign_CPA()$

when received (**FB_ESTIMATE**, $V$, $PA$ , $A_j$)
17. $FC_j(v) \leftarrow V$
18. **if** ( FALB(v)[i] $\geq B$ )
19. $\quad assign_CPA()$

when received (**NEW_SOLUTION**, $PA$)
20. $B_CPA \leftarrow PA$
21. $B \leftarrow PA.cost$

Figure 5: Initialization and message handling procedures of the AFB-BJ Algorithm

The call to **backtrack** is made whenever the current agent cannot find a valid value (i.e. below the bound B). In such a case, the agent calls backtrackTo() to compute to which agent the CPA should be sent, and backtracks the search process (by sending the CPA) back to that agent. If the agent is the first agent (nowhere to backtrack to), the terminate broadcast ends the search process in all agents (line 37). The algorithm then reports that the optimal solution has a cost of $B$, and the full assignment corresponding to this cost is $B_CPA$.

The function **backtrackTo** computes to which agent the CPA should be sent. This is the kernel of the backjumping (BJ) mechanism. It goes over all candidates, from $j-1$ down to 1, looking for the first agent it finds that has a chance of reaching a full assignment with a lower cost than B. FALB(v)[j-1] is a lower bound on the cost of a full assignment extended from CPA[1..j-1], and PC[j]-PC[j-1] is the cost added to that CPA by $A_j$'s assignment. Since $A_j$ picked the lowest cost value in its domain (its domain was ordered in line 11), the addition of these two components

```
procedure assign_CPA:
22. if CPA contains an assignment A_i = w, remove it
23. iterate (from last assigned value) over D_i until the first value satisfying
    v ∈ D_i s.t. CPA.cost + f(v) < B
24. if no such value exists
25.    backtrack()
26. else
27.    assign A_i = v
28.    if CPA is a full assignment
29.       broadcast (NEW_SOLUTION, CPA )
30.       B ← CPA.cost
31.       assign_CPA()
32.    else
33.       send(CPA_MSG, CPA, A_i) to A_{i+1}
34.       forall j > i
35.       send(FB_CPA, A_i, CPA) to A_j

procedure backtrack:
36. if (A_i = A_1)
37.    broadcast(TERMINATE)
38. else
39.    j ← backtrackTo()
40.    remove assignments of A_{j+1}, .., A_i from CPA
41.    send(CPA_MSG, CPA, A_i) to A_j

function backtrackTo:
42. for j = i − 1 downto 1
43.    foreach v ∈ D_i
44.       if ( FALB(v)[j-1] + (PC[j] - PC[j-1]) < B )
45.          return j
46. broadcast(TERMINATE)
```

Figure 6: The assigning and backtracking procedures of the AFB-BJ Algorithm.

produces a more accurate lower bound on the cost of a full assignment extended from CPA[1..j-1]. This can be safely added to the FALB since the it adds a lower bound on the cost increment by an agent for which the FALB did not include a lower bound.

**Example 2** *In the example presented in section 4.1, when $A_3$ computed the FALB(b)[1] it added the past costs of the partial assignments (cost incurred by $A_1$), the local cost of $A_3$, and a lower bound on the cost increment by future agents ($A_4$ and $A_5$). To this sum we can safely add the cost added by $A_2$ if we know that $A_2$ picked its lowest cost assignment.*

This addition helps tighten the FALB and reduce search. If this combined bound is not smaller than $B$, then surely any combination of assignments made by $A_j$ and any following agent could only raise the cost, which is already too high. In case even backjumping back to $A_1$ will not prove helpful, the search process is terminated (line 46).

## 5. Correctness of AFB

In order to prove correctness for $AFB$ two claims must be established. First, that the algorithm terminates and second that when the algorithm terminates its global upper bound $B$ is the cost of the optimal solution. To prove termination one can show that the $AFB$ algorithm never goes into an endless loop. To prove the last statement it is enough to show that the same partial assignment cannot be generated more than once.

**Lemma 1** *The $AFB$ algorithm never generates two identical CPAs.*

Assume by negation that $A_i$ is the highest priority agent (first in the order of assignments) that generates a CPA for the second time. Now lets consider all possible events that immediately preceded this creation.

Case 1 - $A_i$ received a CPA message from a lower priority agent. Let us denote that agent as $A_j$, where $j > i$. When $A_i$ received this message, he executed lines 7-13 (see Figure 3.1). The procedure backtrack in line 14 was not executed since we know $A_i$ generated a CPA, and that procedure would not do so. Therefore line 16 was executed, and the procedure assign_CPA was invoked. $A_i$ executed lines 22-24. Line 25 was not executed since invoking the backtrack procedure could not lead to the creation of the CPA. Therefore, in line 24 a value as described in line 23 was found to exist. Line 23 searches for a value in $A_i$'s remaining value domain, not exploring any value previously attempted for the current set of assignments of higher priority agents. Since we assumed $A_i$ to be the highest priority agent that generates a CPA for the second time, this combination of higher priority assignments did not repeat itself. Therefore, since $A_i$ received the current set of higher priority assignments $A_i$ does not re-pick any local value, and the set of high priority assignments did not repeat itself, therefore $A_i$ cannot pick a value that would generate the same CPA for the second time.

Case 2 - $A_i$ received a CPA message from a higher priority agent. Let us denote that agent as $A_j$, where $j < i$. Since we assumed $A_i$ to be the highest priority agent that generates a CPA for the second time, this combination of higher priority assignments did not repeat itself. Therefore any value $A_i$ would assign next would generate a unique CPA, one which he could not have generated before.

Case 3 - $A_i$ received a CPA message from itself. This cannot be since $A_i$ never sends such a message to itself.

Case 4 - $A_i$ received an FB_ESTIMATE message from $A_j$. $j > i$ since FB_ESTIMATE are only sent in response to FB_CPA messages. Which are only sent (line 34) to agents of lower priority than $A_i$. Since this message caused the creation of a CPA, the condition in line 19 must have been evaluated to be true, and the procedure assign_CPA in line 19 invoked. Similar to case 1, lines 22-24 were executed and line 25 was not. Similar to case 1, a value was found in line 23. This value does not repeat any value previously picked under the current set of higher priority agent assignments. This is the only time the agent received such current set of higher priority agent assignments due to the assumption that $A_i$ is the first to generate a CPA twice.

Case 5 - the procedure init was invoked. This cannot be since no CPAs were previously generated, any CPA generated now must be unique.

No other events could have immediately preceded the creation of the second identical CPA, therefore it is impossible for this event to occur. This completes the proof of the lemma.

Termination follows immediately from Lemma 1.

Next, one needs to prove that upon termination the complete assignment, corresponding to the optimal solution, is in $B_CPA$ (see Figure 3.1). There is only one point of termination for the $AFB$ algorithm, in procedure $backtrack$. So, one needs to prove that during search no partial assignment that can lead to a solution of lower cost than $B$ is discarded. Let us consider all possible cases where an agent discards a CPA, changes a value or skips over a value and let us show that this cannot be. Skipping over or changing a value is only done inside the procedure assign_CPA in lines 22-24. If $v$ is a value that is skipped over, then by the condition itself in line 23 it holds that $CPA.cost + f(v) \geq B$. Since $B \geq B_CPA$, $CPA.cost + f(v) \geq B \geq B_CPA$ and this means that $v$ could not possibly lead to a solution of cost lower than $B_CPA$ at termination. Let us consider all possible cases in which a value is changed. This only occurs inside the procedure assign_CPA. Let us then consider all possible cases in which this procedure is invoked that result in a value change.

Case 1 - invoking assign_CPA from the init procedure (line 4). No solution could be lost since this is the very first assignment performed, no part of the search space is skipped over by this assignment.

Case 2 - invoking assign_CPA from inside the assign_CPA procedure (line 31). This happens when a new best (so far) solution was found. obviously changing the assignment now would not lose this solution since it is saved and broadcasted as the new current solution. It will only be discarded if a better solution is later found.

Case 3 - invoking assign_CPA following a received FB_ESTIMATE message (line 19). The current partial assignment can be safely discarded, knowing that no solution will be lost since the condition in line 18 indicated that the current partial assignment has a lower bound that exceeds the best solution found so far.

Case 4 - invoking assign_CPA following a received CPA_MSG message (line 16) from $A_j$ where $j > i$. This means the CPA returned from a backtrack after fully exploring the current sub-space, and therefore changing the current assignment would not lead to any potential solution lost.

Case 5 - invoking assign_CPA following a received CPA_MSG message (line 16) from $A_j$ where $j < i$. This means that the CPA was received from a higher priority agent. $A_i$ did not yet pick an assignment, so any assignment it will make will not lose out on any potential solutions.

Therefore, any value skipped over and any change to the CPA will not lead to the loss of a potential solution. The only remaining event that may lead to a solution being skipped over is a CPA being discarded. This is done by the time-stamping mechanism and only occurs when the agent knows of the existence of a more up-to-date CPA. That CPA was created because some agent changed its assignment by calling assign_CPA. We showed that in such a case no better solution can be lost, therefore it is safe to discard the CPA.

In conclusion, in any event a value is skipped over or changed or a CPA is discarded, no possible better solution is lost. Therefore at termination, the AFB algorithm reports the best solution possible. This completes the correctness proof of the $AFB$ algorithm. □

In order to prove the correctness of the *AFB-BJ* algorithm we first prove the correctness of the proposed backjumping method and then show that its combination with *AFB* does not violate *AFB*'s correctness which has been proven.

In order to prove the correctness of the backjumping method one need only show that none of the agents' assignments that the algorithm backjumps over, can lead to a solution with a lower cost than the current upper bound. The condition for performing backjumping over an agent $A_j$ (line 44) is that the lower bound on the cost of a full assignment extended from the assignments of $A_1, .., A_{j-1}$ and of the assignment cost of $A_j$ exceeds the global upper bound $B$. Since $A_j$ picked the lowest cost value in its remaining domain (as the domain is ordered), extending the assignments of $A_1, .., A_{j-1}$ must lead to a cost greater or equal to $B$. Therefore, backjumping back to $A_{j-1}$ cannot discard any potentially lower cost solutions. This completes the correctness proof of the *AFB-BJ* backjumping (function *backtrackTo*) method.

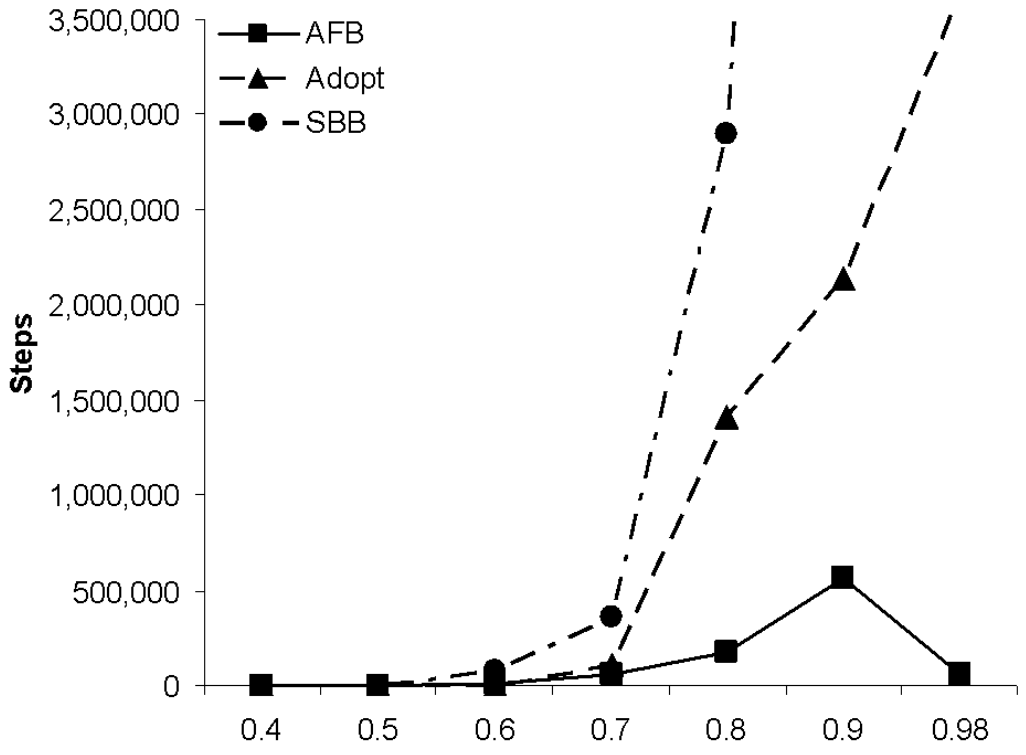

Figure 7: Total non-concurrent computational steps by AFB, ADOPT and SBB on low density ($p_1$=0.4) Max-DisCSP

Assuming the correctness of *AFB*, in order to prove the correctness of the composite algorithm *AFB-BJ* it is enough to prove the consistency of the lower bounds computed by the agents in *AFB-BJ*. The lower bounds computed by *AFB-BJ* include FC, LC and PC as described in section 4. PC is contained in the CPA, and is updated by any agent that receives it and adds an assignment (not shown in the code). LC(v) is computed by the current agent $A_i$ whenever it assigns v as its value assignment. FC$_j$ is computed by $A_j$ in line 5 (in figure 5), and is sent back to $A_i$ in line 6. $A_i$ receives and saves this in line 17. The lower bounds contained inside these vectors are correct because PC was exactly calculated when holding the CPA, LC was exactly calculated by the current agent $A_i$, and the bounds in FC$_j$ are the same bounds computed in *AFB* which were proven to be correct lower bounds for the assignment of $A_j$. The FC$_j$ bounds are accurate and based on the current partial assignment since the timestamp mechanism prevents processing of bounds which are based on an obsolete CPA. Whenever the CPA is altered by some higher priority agent, the previous bounds are cleared (line 10 of figure 5). This completes the correctness proof of $AFB - BJ$. □

## 6. Experimental Evaluation

All experiments were performed on a simulator in which agents are simulated by threads which communicate only through message passing. The Distributed Optimization problems used in all of the presented experiments are random *Max-DisCSPs*. The network of constraints, in each of the experiments, is generated randomly by selecting the probability $p_1$ of a constraint among any pair of variables and the probability $p_2$, for the occurrence of a violation (a non zero cost) among two assignments of values to a constrained pair of variables. Such uniform random constraints networks of $n$ variables, $d$ values in each domain, a constraints density of $p_1$ and tightness $p_2$ are commonly used in experimental evaluations of CSP algorithms (cf. (Prosser, 1996)). *Max-CSP*s are commonly used in experimental evaluations of constraint optimization problems (*COP*s) (Larrosa & Schiex,

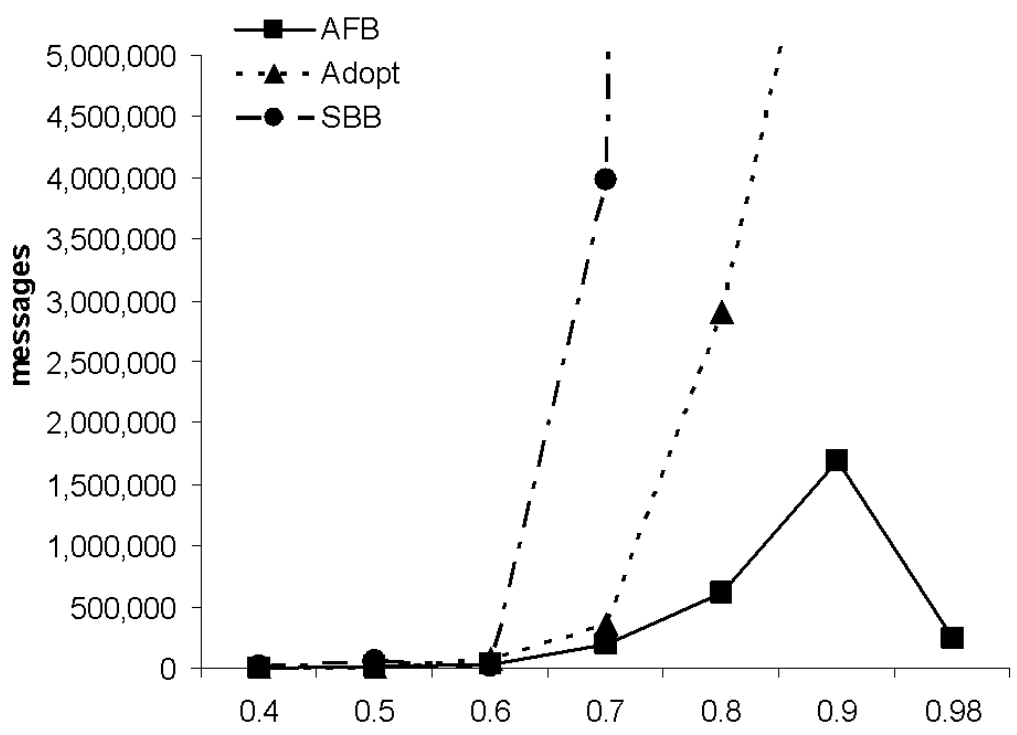

Figure 8: Total number of messages sent by AFB, ADOPT and SBB on low density ($p_1$=0.4) Max-DisCSP

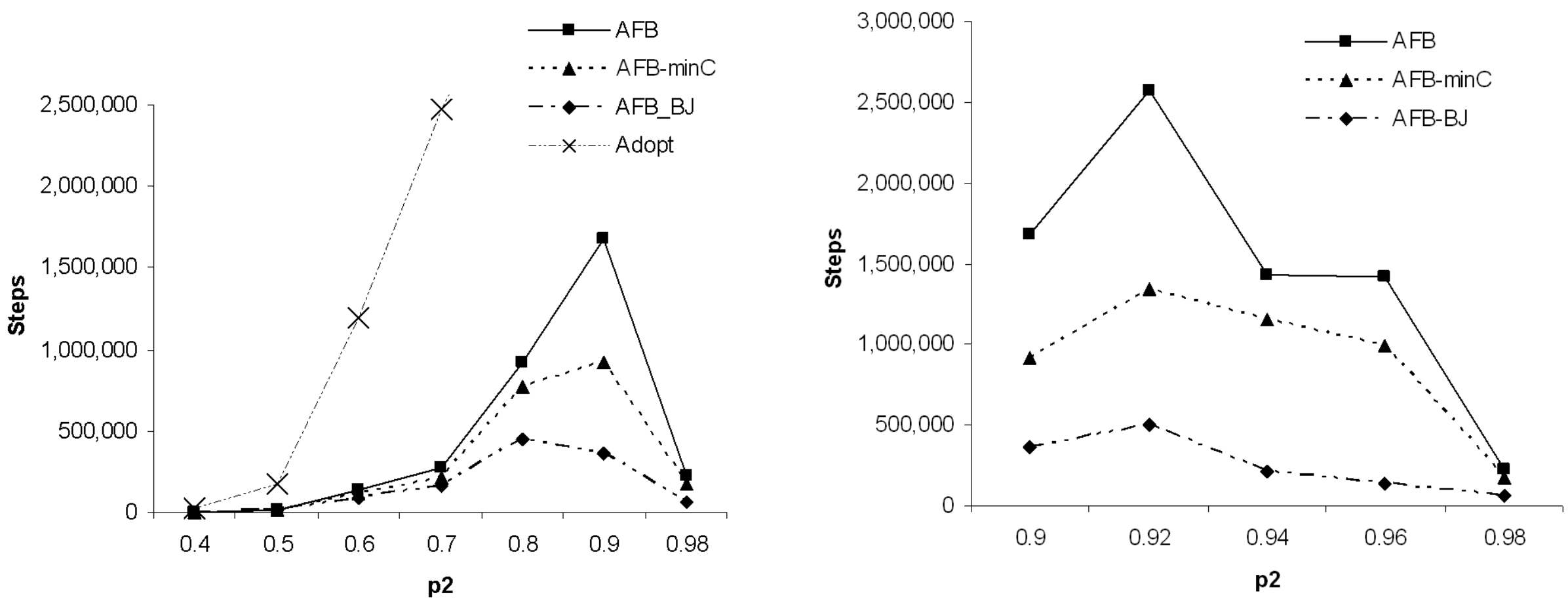

(a) (b)

Figure 9: (a) Number of none-concurrent steps performed by ADOPT, AFB, AFB-minC and AFB-BJ for high density Max-DisCSP ($p_1 = 0.7$). (b) A closer look at $p_2 > 0.9$

2004). Other experimental evaluations of DisCOPs include graph coloring problems (Modi et al., 2005; Zhang et al., 2005), which are a subclass of *Max-DisCSP*.

In order to evaluate the performance of distributed algorithms, two independent measures of performance are used - run time, in the form of non-concurrent steps of computation (Zivan & Meisels, 2006b), and communication load, in the form of the total number of messages sent (Lynch, 1997; Yokoo, 2000a).

In the first set of experiments, the performance of $AFB$ is compared to that of two algorithms. The synchronous $B\&B$ algorithm ($SBB$) (Hirayama & Yokoo, 1997) and the asynchronous distributed optimization algorithm ($ADOPT$) (Modi et al., 2005). Figure 7 presents the average run-time in number of non-concurrent computation steps, on randomly generated Max-DisCSPs with $n = 10$ agents, domain size $d = 10$, and a constraint tightness of $p_1 = 0.4$. Figure 8 compares the same algorithms on the same problems by the total number of messages sent. From these figures it is clear that ADOPT outperforms the basic algorithm SBB, in accordance with the past experimental evaluation of these two algorithms (Modi et al., 2005). It is also clear that AFB outperforms ADOPT by a large margin for tight (high $p_2$) problems. This is true for both measures.

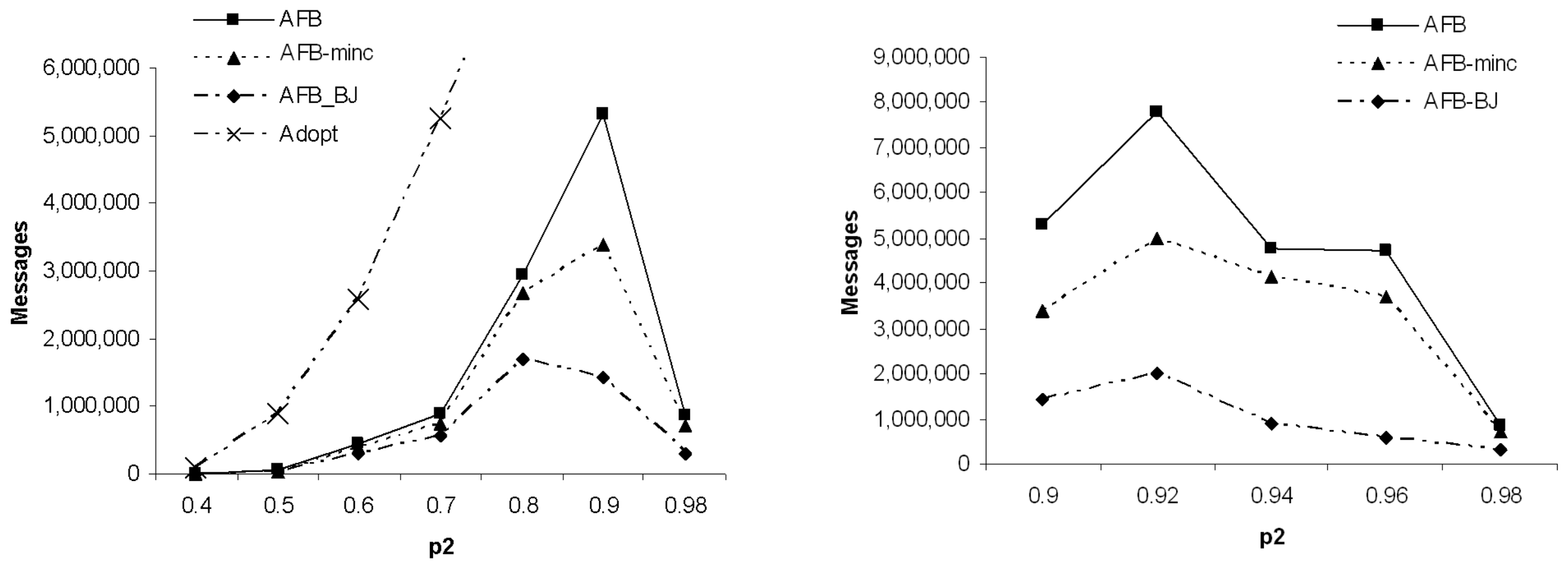

(a) (b)

Figure 10: (a) Number of messages sent by ADOPT, AFB, AFB-minC and AFB-BJ for high density Max-DisCSP ($p_1 = 0.7$). (b) A closer look at $p_2 > 0.9$

The second set of experiments includes the ADOPT algorithm and three versions of the *AFB* algorithm: *AFB*, *AFB-minC* - a variation of *AFB* which includes dynamic ordering of values based on minimal cost (of the current CPA), and *AFB-BJ* which is the composite backjumping and forward-bounding algorithm. *AFB-BJ* uses the same value ordering heuristic as *AFB-minC*. This was selected in order to show that the improved performance of *AFB-BJ* does indeed arise from the backjumping feature and not from the value ordering heuristic.

Figure 9 presents the average run-time in number of non-concurrent computation steps, of all the algorithms: *ADOPT*, *AFB*, *AFB-minC* and *AFB-BJ*, on Max-DisCSPs with $n = 10$ agents, domain size $d = 10$, and a constraint density of $p_1 = 0.7$. Asynchronous optimization (*ADOPT*) is much slower than the standard version of *AFB*. Also clear from this figure, is that the value ordering heuristic greatly improves *AFB*'s performance. The added backjumping improves the performance much further. The RHS of the figure provides a "zoom in" on the section of the graph between $p_2 = 0.9$ and $p_2 = 0.98$. For such tight problems, ADOPT did not terminate in a reasonable amount of time and had to be terminated manually (and thus is missing from the graph).

For tightness values that are higher than $p_2 > 0.9$ *AFB* and its variants demonstrate a "phase transition". This "phase transition" behavior of the *AFB* algorithms is very similar to that of lookahead algorithms on centralized *Max-CSP*s (Larrosa & Meseguer, 1996; Larrosa & Schiex, 2004). Our explanation for this "phase transition" is that problem difficulty increase exponentially with tightness but only up to some point. When the problem becomes over-constrained such that many combinations produce the highest cost possible all these combinations are in fact equal in quality, and can be easily pruned by an intelligent search.

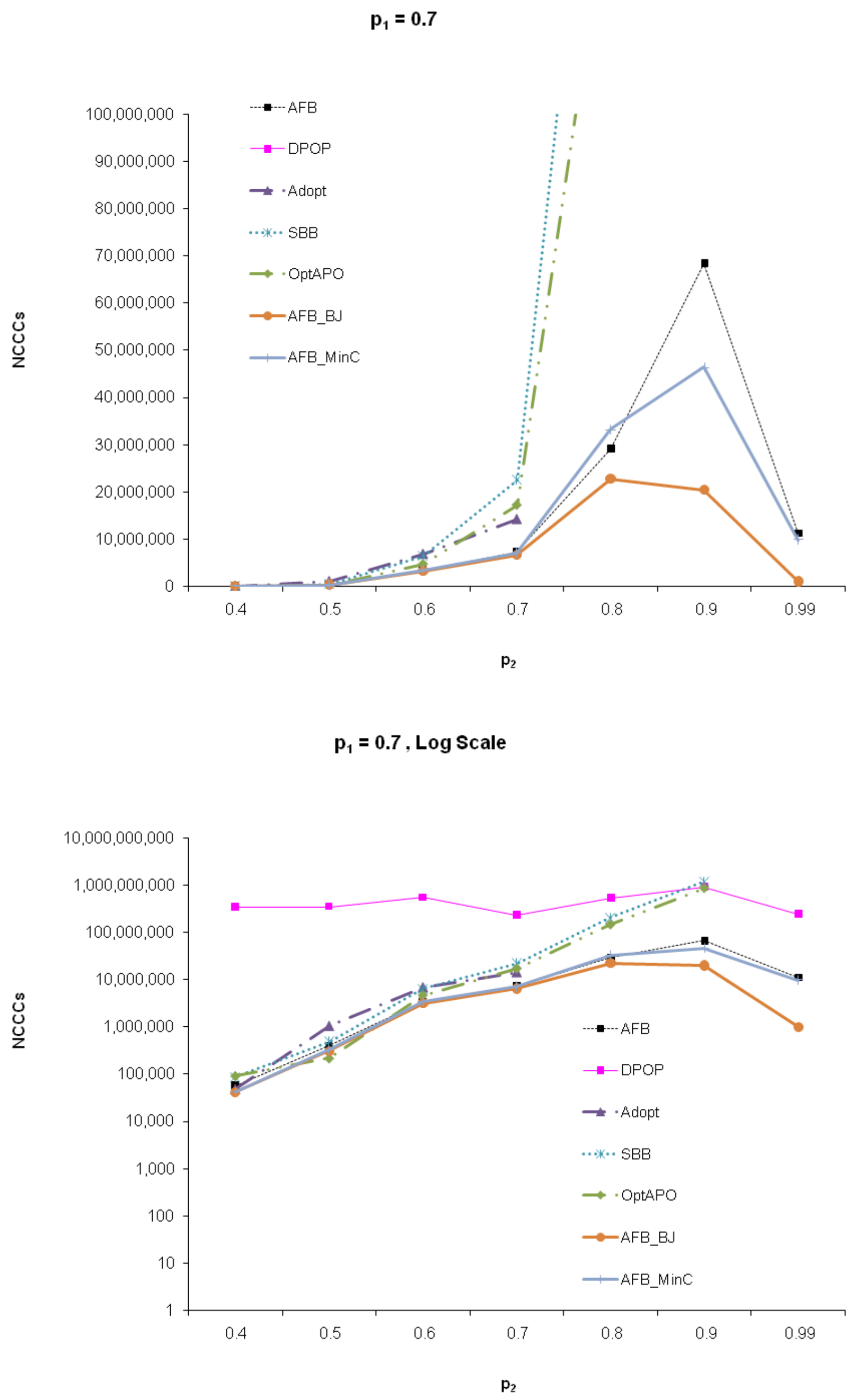

Figure 11: Number of Non-Concurrent Constraint Checks (NCCCs) performed by several DisCOP solvers for high density Max-DisCSP ($p_1 = 0.7$) in both linear scale (top) and logarithmic scale (bottom)

Figure 10 presents the total number of messages sent by each of the algorithms. The results of this measurement closely match the results of run-time, as measured by non-concurrent steps.

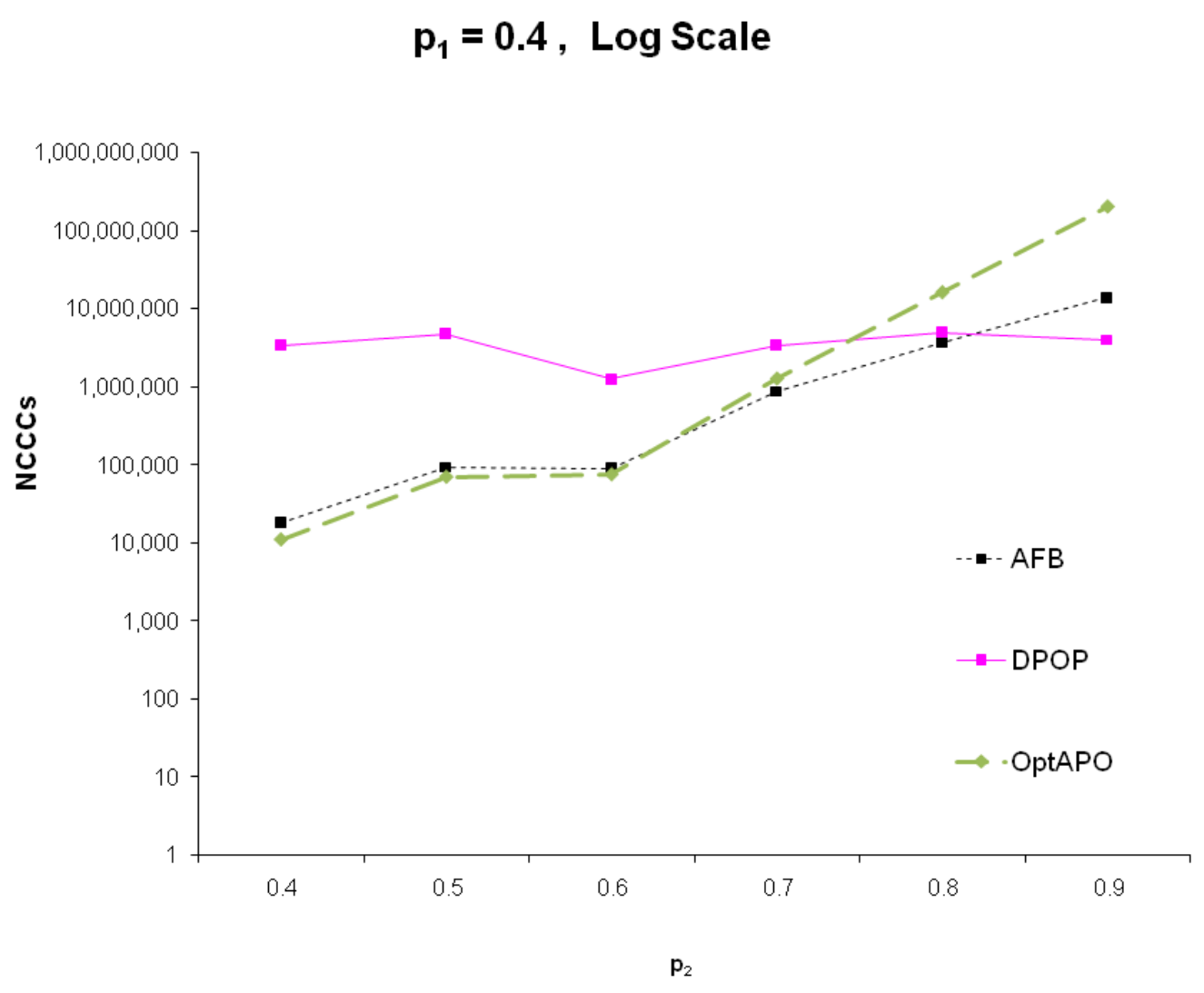

Figure 12: Number of Non-Concurrent Constraint Checks (NCCCs) performed by several DisCOP solvers for low density MaxDisCSP ($p_1 = 0.4$) in logarithmic scale

We can see that ADOPT has an exponentially rapid growth of messages. The explanation for this growth is simple. Following each message an agent receives in ADOPT, several VALUE messages are sent to lower priority agents, and a single COST message is sent to a higher priority agent (Modi et al., 2005). On the average, at least two messages are sent for every message received, therefore the total number of messages in the system increases exponentially over time.

The third batch of experiments, includes a comparison with two additional DisCOP solvers - DPOP (Petcu & Faltings, 2005a) and OptAPO (Mailler & Lesser, 2004). DPOP performs only a linear number of computational steps, but each step performs an exponential number of computations. The number of messages in DPOP is linear ($2n$) in the number of agents. Similar to ADOPT, DPOP also uses a pseudo-tree ordering of the agents and so we use the same ordering for both algorithms. OptAPO performs a partial centralization of the problem, and has agents that solve a part of the problem they are in charge of. Therefore, for both algorithms, evaluation measures that use the number of (non-concurrent) *computational steps* are inappropriate, since the steps can be exponentially time consuming. For this reason, the performance of all algorithms must be evaluated by a different metric. The canonical choice is the number of non-concurrent constraint checks ($NCCCs$). This implementation independent measure includes the computations performed within every single step (Zivan & Meisels, 2006b, 2006a, 2006). The number of messages sent is also not a good measure in this case, since DPOP sends out exponentially large messages (but only a linear number of them) while the other algorithms send out an exponential amount of messages but of only linear size. Thus we only present the results using the $NCCCs$ metric. We repeat the experimental setup of the previous experiment on randomly generated problems, and report the total number of non-concurrent constraint checks (NCCCs) in figure 11. The results are presented in both logarithmic and linear scales.

In this experiment OptAPO, SBB and ADOPT did not terminate in a reasonable time on some of the harder problem instances and are therefore partially absent in the graphs. The computation in DPOP is composed of each agent sending out a message containing its subtree's optimal cost for every possible combination of higher priority constrained agents. For a given constraint density the size of the message each agent sends would not be effected by changing the constraint tightness. Therefore, the computation performed by each agent is unaffected by changing the constraint tightness ($p_2$). DPOP's run time is expected to remain roughly the same for all tightness values in our experiment. For problems with a low constraint tightness DPOP's performance is poor when compared to the rest of the algorithms. However, as problem tightness increases the gap between DPOP's run time and the rest of the algorithms narrows, until at $p_2 = 0.9$ DPOP and OptAPO and SBB have roughly the same run time. At $p_2 = 0.99$ DPOP outperforms ADOPT, OptAPO and SBB (which did not terminate). AFB and its variants outperform DPOP for the whole range of constraint tightness by orders of magnitude. OptAPO appears to perform only slightly better than SBB and AFB clearly outperforms it by orders of magnitude. AFB and its variations produce the same "phase transition" as reported in previous experiments, and $AFB - BJ$ comes out as the best performing algorithm for solving random DisCOPs.

The results for a similar experiment in low density ($p_1 = 0.4$) Max-DisCSPs are presented in figure 12 (notice the logarithmic scale). As in high density problems, DPOP performance is unaffected by the problem tightness, producing roughly similar results for all tightness values. At low tightness values, OptAPO and AFB are vastly superior to DPOP while OptAPO slightly outperforms AFB. As tightness increases, OptAPO increases exponentially in run-time to become the worst performing algorithm. AFB outperforms DPOP at all tightness values except at $p_2 = 0.9$.

## 7. Conclusions

The Asynchronous Forward-Bounding algorithm ($AFB$) uses asynchronous and concurrent constraint propagation on top of the distributed Branch and Bound scheme. In its forward-bounding protocol $AFB$ maintains local consistency, and prevents exploration of "dead-ends" of the search-space. The run-time and network load of AFB were evaluated by an asynchronous simulator on randomly generated $Max - DisCSPs$. The results of this evaluation revealed a phase-transition in $AFB$'s performance, as the tightness of the problems increased beyond some point. No other $DisCOP$ solver was reported to display such a behavior. A similar phase-transition was previously reported for centralized $COP$ solvers, as part of the work of Larrosa et. al. (Larrosa & Meseguer, 1996; Larrosa & Schiex, 2004). The phase-transition observed there is reported to occur only by $COP$ solvers, that enforce a strong enough form of local consistency (Larrosa & Meseguer, 1996; Larrosa & Schiex, 2004). We therefore attribute this behavior of AFB to its *concurrent enforcement of local consistency*.

$AFB$ can be extended. One extension is to include a value ordering heuristic. A good ordering heuristic is the minimum-cost heuristic, where values with lower cost due to assignments of higher priority agents are selected first. We named this version of the algorithm *AFB-minC*. In the experiments, the use of this heuristic substantially improved the performance of $AFB$.

A further extension of $AFB$ enhanced it with a backjumping mechanism. By adding a small amount of information to the bounding messages, agents which detect that the lower bound of the current partial assignment is too large (i.e. the state is inconsistent and backtracking is required) are now able to check whether backtracking to the previous agent will indeed help to reduce the lower bound so that the resulting partial assignment is consistent. Otherwise, the search process backtracks even further. The resulting algorithm, *AFB-BJ*, performs significantly better than the other versions of AFB. By comparing *AFB-minC* and *AFB-BJ*, it was shown that the backjumping does indeed affect performance, and the improvement over standard $AFB$ is not only a result of the addition of the ordering heuristic.

The $AFB$ algorithm was compared to two algorithms that are based on the branch & bound mechanism in its distributed form - ADOPT and SBB (Yokoo, 2000b; Modi et al., 2005). The experimental evaluation clearly demonstrates a substantial difference in performance between the algorithms. Asynchronous distributed optimization ($ADOPT$) outperforms $SBB$, but $AFB$ outperforms $ADOPT$ by a large margin in both measures of performance. To the best of our knowledge this is the only evaluation of $ADOPT$ on increasingly tighter problems. Other experimental evaluations measured $ADOPT$'s scalability (by increasing the number of variables) and not by increasing the difficulty (tightness) of problems of a fixed size. The exponential growth of the number of messages in $ADOPT$ is also apparent in Figures 8 and 10(a). Outperforming $AFB$ are the two extended versions of $AFB$, ***AFB-minC*** and ***AFB-BJ***, with ***AFB-BJ*** having the best performance. The proposed value ordering heuristic improves performance, and when adding the backjumping mechanism on top of that, performance is even further enhanced.

Although $AFB$ and $ADOPT$ perform concurrent computation the nature of concurrency used by them is very different. Concurrency in $ADOPT$ is achieved by performing asynchronous assignments. In such an algorithm each agent picks its value assignment and is free to change it at any time. Multiple agents may change their assignments concurrently. Asynchronous assignments introduce some degree of uncertainty with regard to the consistency of the current partial assignment as known to an agent. In fact, there are scenarios in which an agent may base its computation on an inconsistent partial assignment, which is a combination of assignments performed by higher priority agents that are not aware of each other's most-up-to-date assignment.

Two algorithms that were used for comparisons with $AFB$ - $ADOPT$ and $DPOP$ - use the pseudo-tree ordering of agents, which allows independent subproblems to be solved concurrently. A good pseudo-tree ordering can be problematic to find (it is NP-hard to find the optimal ordering), and sometimes even the best ordering is not good enough, due to the structure of the specific problem. Overall, these orderings become less useful when dealing with problems with high constraint density.

In order to further evaluate the performance of AFB, it was compared and tested against two additional DisCOP algorithms. Both DPOP and OptAPO do not use branch and bound to find an optimal solution. The DPOP algorithm delivers all possible partial assignments up the pseudo-tree and performs an exponential number of constraints checks in two passes over the pseudo-tree (Petcu & Faltings, 2005a). OptAPO partitions the DisCOP into sub-problems, each solved by a mediator of that sub-problem (Mailler & Lesser, 2004). The performance of these algorithms is expected to be different than algorithms that use branch & bound search. In fact, the performance of DPOP on randomly generated DisCOPs is independent of the tightness of the problems. The results of extensive empirical evaluations of all algorithms on random DisCOPs are described in section 6 and are conclusive. The AFB algorithm is the best performing DisCOP algorithm on randomly generated DisCOPs in both measures of performance. It performs less non-concurrent constraints checks and it sends a smaller number of messages.

In essence, the idea behind $AFB$ can be summed up as follows - run a sequential assignment optimization process and concurrently run in parallel many additional processes that check the consistency of the partial assignment. The main search process is slow. At any point in time only one agent holds the current partial assignment in order to extend it. Concurrency is achieved via the forward bounding, which is performed concurrently.

The results of the experimental evaluation show that adding concurrent maintenance of bounds to a sequential assignment process results in an efficient optimization algorithm ($AFB$). This algorithm outperforms all other concurrent algorithms on the hard instances of random DisCOPs.